\documentclass[runningheads]{llncs}
\usepackage[T1]{fontenc}
\usepackage{graphicx}
\usepackage{booktabs}
\usepackage{amsmath}
\usepackage{amsfonts}
\usepackage{multirow}
\usepackage{array} 
\usepackage{makecell} 
\usepackage{relsize} 
\usepackage{caption}
\begin{document}
\title{
Controllable and Efficient Multi-Class Pathology Nuclei Data Augmentation using Text-Conditioned Diffusion Models
}
\titlerunning{Text-Conditioned Diffusion Models for Multi-Class Nuclei Data Aug.}
%
\author{Hyun-Jic Oh \and Won-Ki Jeong} 
\authorrunning{H.J. Oh and W.K. Jeong}
%
%
\institute{
College of Informatics, Department of Computer Science and Engineering, \\
Korea University, Seoul, South Korea
\\
\email{wkjeong@korea.ac.kr}
}

%
\maketitle              
\begin{abstract}

In the field of computational pathology, deep learning algorithms have made significant progress in tasks such as nuclei segmentation and classification. 
However, the potential of these advanced methods is limited by the lack of available labeled data.
Although image synthesis via 
recent generative models has been actively explored to address this challenge, 
%
%
%
existing works have barely addressed label augmentation and are mostly limited to single-class and unconditional label generation.
In this paper, we introduce a novel two-stage framework for multi-class nuclei data augmentation using text-conditional diffusion models.
In the first stage, we innovate nuclei label synthesis by generating multi-class semantic labels and corresponding instance maps through a joint diffusion model conditioned by text prompts that specify the label structure information.
In the second stage, we utilize a semantic and text-conditional latent diffusion model to efficiently generate high-quality pathology images that align with the generated nuclei label images.
We demonstrate the effectiveness of our method on large and diverse pathology nuclei datasets, with evaluations including qualitative and quantitative analyses, as well as assessments of downstream tasks.

\end{abstract}

\section{Introduction}


In digital pathology, nuclei segmentation and classification are essential tasks for accurate disease diagnosis and prognosis. 
These processes allow for the quantitative analysis of complex nucleus properties such as size, shape, and distribution~\cite{alom2022microscopic}. 
Recent advances in deep learning-based methods have significantly improved these processes~\cite{graham2019hover,ilyas2022tsfd}.
However, the limited availability of labeled data remains a challenge, restricting the predictive capabilities of these algorithms. The time-consuming nature of pathology image acquisition and labeling further worsens this problem. 

To overcome these challenges, data synthesis using deep generative models has been emerging as a promising solution.
While generative adversarial networks (GANs)~\cite{spade2019,pix2pixHD2018} have been widely employed for image generation~\cite{gong2021style,wang2023sian}, recent studies show that diffusion models outperform GANs in natural images~\cite{diffusion_beat_gans2021,ddpm2020,sdm2022}, making them a new standard method in many image synthesis tasks. 
%
These approaches aim to address the efficiency of data collection and enhance dataset diversity.

Although generative models have been successful in synthesizing natural images, we have observed several challenges in adopting this method for pathology image synthesis. 
For example, SDM~\cite{sdm2022} utilizes semantic labels as conditions for image synthesis, which is later extended to pathology nuclei image synthesis tasks~\cite{diffmix2023,nasdm2023,nudiff2023}. 
Despite their ability to produce high-quality images, pixel-based diffusion models are less practical for data augmentation due to the time and computational cost required for training and inference procedures.
Moreover, existing methods have mainly focused on image synthesis rather than realistic label augmentation; many nuclei data augmentation methods rely on a simple geometric alteration, such as random perturbation of nuclei positions~\cite{diffmix2023}, copy and pasting~\cite{doan2022gradmix}, or deformation~\cite{gong2021style}, without consideration to the spatial context of the labels. 
Yu~\textit{et al.}~\cite{nudiff2023} employed an unconditional diffusion model~\cite{ddpm2020} that synthesizes nuclei labels with distance transform masks for both dense pixel semantic and instance label generation. 
However, their approach lacks scalability for multi-class data, and the standard Gaussian diffusion model is not optimal for multi-class semantic data synthesis. 
Instead, Hoogeboom~\textit{et al.}~\cite{hoogeboom2021argmax} suggested utilizing categorical distribution for semantic data synthesis, and Park~\textit{et al.}~\cite{park_gcdp2023} introduced a joint Gaussian-categorical diffusion model that generates image-label pair synthesis.
However, we observed that unconditional diffusion models tend to replicate frequently observed structures in the training data, resulting in biased class distribution, which is especially severe in pathology nuclei label synthesis. 

In this paper, we propose a novel two-stage multi-class pathology nuclei data augmentation framework addressing the above issues. 
First, we develop a text-conditional joint diffusion model for spatially controllable multi-class nuclei label synthesis, inspired by 
the text-conditional generation model~\cite{saharia2205photorealistic} that utilizes text-image alignment with target-related text prompts to generate fine-grained images. 
To achieve this, we utilize a pre-trained text encoder that is specifically designed for pathology images~\cite{pathldm2024}.
%
Second, we formulate a semantic and text conditional latent diffusion model (LDM) based on the pretrained LDMs (e.g., Zhang~\textit{et al.}~\cite{zhang2023_controlnet}) and fine-tune it for pathology nuclei image synthesis, which significantly enhances performance and efficiency in the sampling process. 
We demonstrate the effectiveness of our method across various datasets stained with different modalities, conducting comprehensive evaluations that assess both the quality of synthesized data and the performance of downstream tasks. 
Our contributions can be summarized as follows:
\begin{itemize}
    \item 
    We introduce a novel two-stage framework for pathology nuclei data augmentation consisting of multi-class label synthesis and high-quality realistic image synthesis. 
    We pioneer a multi-class dense pixel-by-pixel label generation strategy and tailor the pretrained LDMs to synthesize pathology images corresponding to the generated nuclei label images, ensuring both image quality and computational efficiency. 
    \item 
    We propose text conditioning for detailed guidance of the label synthesis process to mitigate the inherent bias of unconditional models, which typically favor generating labels with nucleus proportions frequently observed in the training data. 
    Our diffusion-based method implicitly learns the spatial layout of nuclei while text conditioning allows fine control of class distribution in the synthesized image, which effectively avoids data imbalance problems. 
    \item 
    We validate our methodology on large public datasets with a variety of staining modalities, and its effectiveness has been demonstrated through comprehensive qualitative and quantitative analysis, along with evaluation of downstream tasks.
\end{itemize}

\begin{figure}[t!]
    \centering
    \includegraphics[width=12.1cm, keepaspectratio]{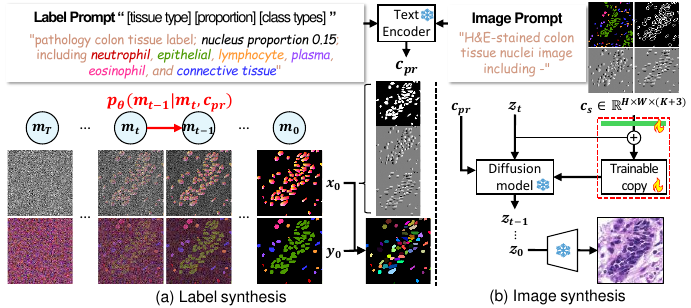}
    \caption{Overview of the proposed two-stage data synthesis framework, consisting of label and image synthesis steps. The framework utilize spatial text condition $c_{pr}$ for label synthesis (a). For image synthesis (b), we fine-tune the pretrained latent diffusion model with semantic condition $c_s$ and text condition $c_{pr}$.
    }
    \label{fig:framework}
\end{figure}




\section{Method}
In this section, we introduce a novel two-stage data augmentation framework for multi-class pathology nuclei data using diffusion models. 
We provide background on diffusion models (Sec.~\ref{sec:diffusion_processes}), followed by detailed descriptions of our approach to text conditional label synthesis (Sec.~\ref{sec:label_synthesis}) and LDM-based semantic and text conditional image synthesis (Sec.~\ref{sec:image_synthesis}). 
The overview of the framework is illustrated in Fig.~\ref{fig:framework}.

\subsection{Background: Diffusion Models}
\label{sec:diffusion_processes}


Diffusion models~\cite{ddpm2020} synthesize data from a learned data distribution by converting noise, operating on a forward and backward diffusion processes. 
In the forward process $q(\mathbf{x}_t|\mathbf{x}_{t-1})$, noise $\epsilon_t$ is incrementally added to data $\mathbf{x}_0$ over timesteps $t\in\{1,...,T\}$ according to a predefined noise schedule $\beta_t$, resulting in a pure noise $\mathbf{x}_T$ that aligns with a specific distribution.
Then, the model $\epsilon_\theta$ learns to predict the noise by the backward process $p_{\theta}(\mathbf{x}_{t-1}|\mathbf{x}_t)$ to reconstruct $\mathbf{x}_0$. 
Standard diffusion models generally employ a Gaussian noise for ordinal data synthesis such as images.
The diffusion process is then defined as:
\begin{equation}
    q(\mathbf{x}_t | \mathbf{x}_{t-1}) := \mathcal{N}(\mathbf{x}_t; \sqrt{1 - \beta_{t}}\mathbf{x}_{t-1}, \beta_t\mathbf{I}),
\end{equation}
\begin{equation}
    p_{\theta}(\mathbf{x}_{t-1} | x_t) := \mathcal{N}(\mathbf{x}_{t-1}; \mu_{\theta}(\mathbf{x}_t), \sigma_t^2\mathbf{I}).
\end{equation}
The training objective is then defined as:
\begin{equation}
    \mathbb{E}_{x,\epsilon\sim\mathcal{N}(0,1),t} \left[ \left\| \epsilon - \epsilon_{\theta}(\mathbf{x}_t, t) \right\|^2_2 \right].
\end{equation}

For synthesizing categorical data $\mathbf{y}$ such as semantic labels, a categorical diffusion model is employed, using a categorical distribution $\mathcal{C}$ as defined by Hoogeboom~\textit{et al.}~\cite{hoogeboom2021argmax}:
\begin{equation}
    q(\mathbf{y}_t|\mathbf{y}_{t-1}) := \mathcal{C}(\mathbf{y}_t; (1-\beta_t)\mathbf{y}_{t-1} + \beta_t/K),
\end{equation}
\begin{equation}
    p_\theta(\mathbf{y}_{t-1}|\mathbf{y}_t) := \mathcal{C}(\mathbf{y}_{t-1}; \Theta_\theta(\mathbf{y}_t)),
\end{equation}
where $K$ represents the number of categories and $\Theta$ is the probability mass function of $\mathcal{C}$.

For pair-wise synthesis of ordinal and categorical data, a Gaussian-categorical diffusion model~\cite{park_gcdp2023} formulates a joint distribution of Gaussian and categorical distributions $\mathcal{NC}$ as:
\begin{equation}
    p_\theta(\mathbf{m}_{t-1}|\mathbf{m}_t) := \mathcal{NC}(\mathbf{m}_{t-1}; \mu_\theta(\mathbf{m}_t), \Sigma_\theta(\mathbf{m}_t), \Theta_\theta(\mathbf{m}_t)),
\label{eq:joint_dp}
\end{equation}
where $\mathbf{m}$ indicates the data pair $(\mathbf{x},\mathbf{y})$.

\subsection{Text Conditional Multi-class Label Synthesis}
\label{sec:label_synthesis} 

\begin{figure}[t!]    
    \centering
    \includegraphics[width=12.2cm, keepaspectratio]{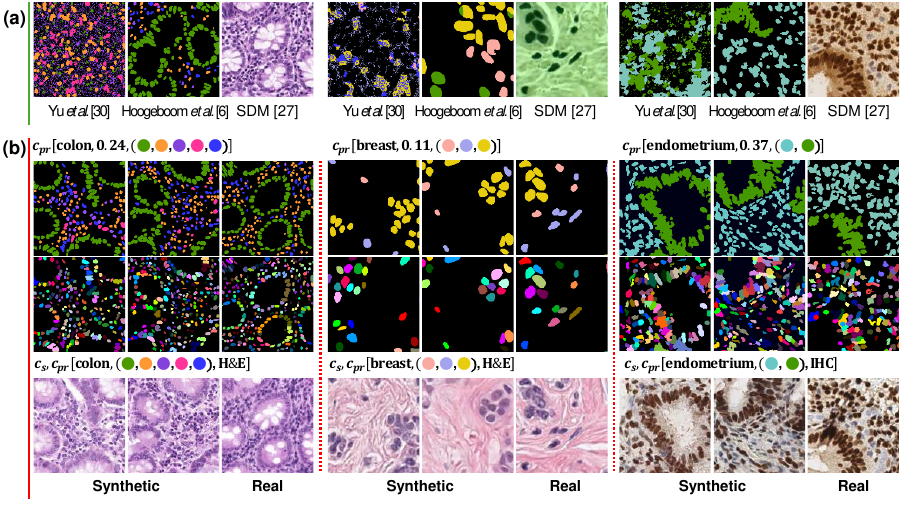}
    \caption{
    Example of generated data using text conditions $c_{pr}$ (b). (a) shows the synthetic data by related works.
    }
    \label{fig:overview}
\end{figure}

Synthetic nuclei label requires an corresponding instance label to support advanced nuclei analysis algorithms.
Considering the difficulty of nucleus instance separation because of clustered nucleus, 
we formulate a text conditional joint diffusion model that simultaneously generates semantic label with structure map for nuclei separation.
First, we define the semantic label as one-hot shaped vector $y \in \mathbb{R}^{H\times W \times K}$ with categorical diffusion process, where $K$ indicates the number of classes. 
Additionally, we define a 3-channel structure map $x \in \mathbb{R}^{H\times W \times 3}$ as the concatenation of a binarized semantic label with x- and y-directional distance transform maps inspired by~\cite{nudiff2023} with Gaussian diffusion process (see $x_0$ in Fig.~\ref{fig:framework}a). 
We define a data pair $m_0=(x_0,y_0) \in \mathbb{R}^{H \times W \times 6}$ by concatenating $x_0$ and $y_0$ along channel dimension, while $y_0$ is embedded into 3-channel via learnable parameters. 
Then, the joint diffusion model is trained to generate a data pair $m_0$ as Eq.~\ref{eq:joint_dp}.

Furthermore, we incorporate a pre-trained text encoder in the pathology domain, PLIP~\cite{pathldm2024}, encoding text prompts depicting tissue type, nuclei proportions, and class types while freezing the parameters of the text encoder. 
This allows fine control over the label generation process through the alignment of text and image (see Fig.~\ref{fig:overview}), preventing the learning bias in the distribution of the training data that unconditional models have (see Fig.~\ref{fig:graph}a).
Conditioned by text embedding $c_{pr}$, the joint diffusion model is trained through the backward diffusion process $p_\theta(m_{t-1}|m_t, c_{pr})$ as depicted in Fig.~\ref{fig:framework}a, enabling more sophisticated label generation using text-image alignments.
Lastly, we create instance label using synthesized semantic label $y_0$ and distance masks $x_0$ by marker-controlled watershed algorithm~\cite{yang2006nuclei}. 

\subsection{Conditional Image Synthesis with Pretrained LDM}
\label{sec:image_synthesis}


Latent diffusion model (LDM)~\cite{rombach2022_stablediffusion} compresses an input image into lower dimensional vector $z$ and performs the diffusion process within the latent space, lowering the cost required for training and inference of diffusion models.
The generated sample $z_0$ is then reconstructed into a high resolution image in the pixel space.
Since training an LDM from scratch consumes much resources, we fine-tune a text conditional LDM for pathology, PathLDM~\cite{pathldm2024}, to leverage a rich domain-specific knowledge without changing the pretrained weights.
To synthesize the images reflecting semantic conditions, we introduce a scalable component to influence the generation process.
Inspired by~\cite{zhang2023_controlnet}, the scalable component is a trainable copy of the diffusion model that is connected to the original model via zero-initialized convolutional layers to prevent adversely affecting noise to the network's parameters.
We only train this scalable component with semantic condition $c_s \in \mathbb{R}^{H\times W \times (K+3)}$, which is a combination of semantic labels, distance transform maps, and instance edge map.
As illustrated in Fig.~\ref{fig:framework}b, the semantic condition passes through learnable embedding layer, and computed with $z_t$.
The training objective is then formulated as:
\begin{equation}
    \mathbb{E}_{z,\epsilon\sim\mathcal{N}(0,1),t,c_s,c_{pr}} \left[ \left\| \epsilon - \epsilon_{\theta}(z_t, c_s, c_{pr}, t) \right\|^2_2 \right],
\end{equation}
where $c_{pr}$ represents the text embedding.

\section{Experiment}

\subsection{Datasets}

We utilize three nuclei segmentation datasets: the first two, Lizard~\cite{graham2021lizard} and PanNuke~\cite{gamper2020pannuke}, were stained with hematoxylin and eosin (H\&E), and the third, EndoNuke~\cite{naumov2022endonuke}, was stained using immunohistochemistry (IHC) techniques. 
Lizard is the largest publicly available pathology nuclei dataset, consisting of colon tissue samples. It is derived from six different data sources and contains 495,179 nuclei, 
representing collections from multiple institutions. We pre-process the image samples into 13,064 256$\times$256 sized patches as done by~\cite{nasdm2023}.
PanNuke is a pan-cancer nuclei dataset, incorporating 19 different tissue types. It features 205,343 nuclei across 7,901 patches, each 256$\times$256 pixels in size at 40x higher magnification.
Finally, EndoNuke is a endometrial tissue dataset, including 210,419 nuclei labels. 
These images have been resized and organized into a set of 1,780 patches, each 256$\times$256 pixels in size.
For training the generative model, we divided each dataset into training and test sets with these ratios: 80:20 for PanNuke, 85:15 for EndoNuke, and the NASDM~\cite{nasdm2023} split for the Lizard dataset.

\subsection{Implementation details}
We use a single A6000 GPU with batch size 6 for the label synthesis model and a single RTX4090 GPU with batch size 16 for the image synthesis model.
We set the timestep $T=1000$ for training and $T=100$ for DDIM-based sampling~\cite{song2020denoising_ddim} for both models. 
%
For image and label synthesis, we utilize distinct text prompts. The label synthesis prompts are based on the ground truth labels and data descriptions. These prompts define the tissue type (e.g., colon), nucleus proportion that is ratio of pixels representing nuclei in the label map, and identify the types of nuclei present. Similarly, for image generation, the prompts include both tissue type and nuclei type, as used for label synthesis, and additionally specify the staining method (e.g., H\&E or IHC).
%
For the label synthesis model, we utilized the U-Net model from~\cite{nichol2021improved} and modified input/output channels as data pairs and followed text conditioning as done by Imagen~\cite{saharia2205photorealistic}.

\begin{table}[t!]
\caption{
Quantitative results on generated labels using FSD metric and inference time per sample in second. 
[$\cdot$] is type of text conditions: $Tis$, $NP$, and $C$ indicates tissue type, nucleus proportion, and class types, respectively.
}
\centering
\begin{tabular}
{
l|c|c|c|c
}
\toprule\specialrule{0.05pt}{0pt}{0pt}
\textbf{Method} & \textbf{Lizard} & \textbf{PanNuke} & \textbf{EndoNuke} & \textbf{Time (sec)}  \\ 
\hline\hline
Yu~\textit{et al.}~\cite{nudiff2023}             & 1018.19 & 1130.74  &  940.41 & 122.14                                        \\ \hline
Hoogeboom~\textit{et al.}~\cite{hoogeboom2021argmax} &   36.20 &  438.19  &  141.49 &  28.73                                        \\ \hline\hline
Ours (Uncond.)                                   &   12.00 &   41.28  &  190.81 &      -                                        \\ 
+ $c_{pr}[Tis]$                                  &       - &    9.11  &       - &      -                                         \\ 
+ $c_{pr}[Tis, NP]$                              &  4.40   &    4.45  &   60.42 &     -           \\ 
+ $c_{pr}[Tis, NP, C]$                            &  \textbf{4.10}  &  \textbf{4.31}   &  \textbf{15.11} & \textbf{17.85}  \\ 

\toprule 

\end{tabular}
\label{tab:label_synthesis}
\end{table}




\begin{figure}[t]
    \centering
    \includegraphics[width=11.3cm, keepaspectratio]{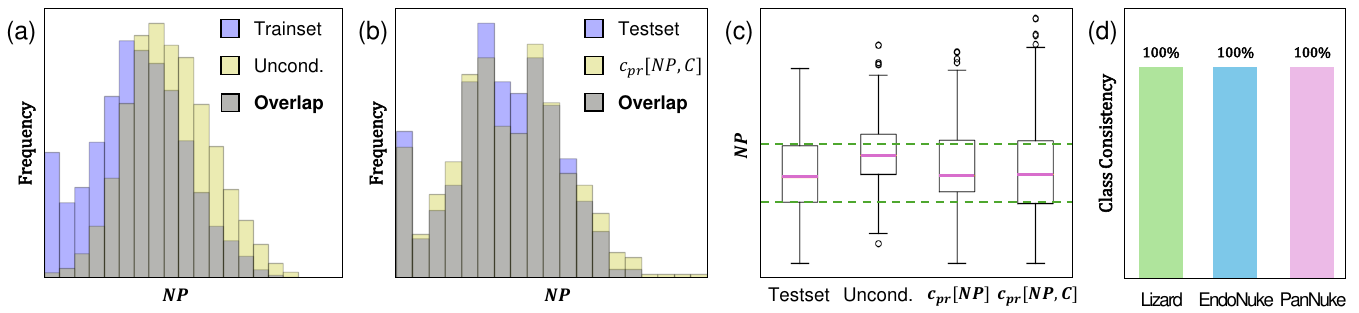}
    \caption{
    Graphical analysis.
    The unconditional model tends to replicate the high-frequency labels from the training data set (a), while the conditional model can generate labels with specific nucleus proportions (b).
    (c) shows that the nucleus proportion and class type conditions are effective for synthesizing the target label set.
    We also observe that it is 100\% consistent for the given class types as (d).
    }
    \label{fig:graph}
\end{figure}

\begin{table}[t]
\caption{
Quantitative comparison of generated image quality to semantic condition-based generative models with FID and IS metrics, and required time to generate a sample in second.
}
\centering
\begin{tabular}
{
l|cc|cc|cc|c
}
\toprule\specialrule{0.05pt}{0pt}{0pt}
\multirow{2}{*}{\textbf{Method}} & \multicolumn{2}{c|}{\textbf{Lizard}} & \multicolumn{2}{c|}{\textbf{PanNuke}} & \multicolumn{2}{c|}{\textbf{EndoNuke}} & \multirow{2}{*}{\textbf{Time (sec)}} \\ \cline{2-7} 
\multicolumn{1}{c|}{} & FID$\downarrow$ & IS$\uparrow$ & FID$\downarrow$ & IS $\uparrow$ & FID$\downarrow$ & IS $\uparrow$ & \multicolumn{1}{c}{} \\ \hline\hline
pix2pixHD~\cite{pix2pixHD2018} & 99.97 & 1.59 & 100.36 & 2.82 &  95.37 & 1.95 & - \\  \hline    
SPADE~\cite{spade2019} & 60.93 & 1.71 &  \underline{71.03} & 3.35 & \underline{68.02} & 1.63 & - \\  \hline    
SDM~\cite{sdm2022} & \underline{48.16} & \textbf{2.45} & 95.44 & \textbf{3.92} & 91.42 & \textbf{2.24} & 180 \\  \hline
Ours & \textbf{37.69} & \underline{2.36} &  \textbf{50.29} & \underline{3.59} & \textbf{57.18} & \underline{2.14} &  5.35 \\ \toprule
\end{tabular}
\label{tab:image_synthesis}

\end{table}

\begin{table}[]
\caption{
Downstream segmentation and classification performance. 
The best results are in \textbf{bold} and the second best are \underline{underlined}.
}
\centering
\begin{tabular}{
l|ccc|ccc|ccc
}
\toprule\specialrule{0.05pt}{0pt}{0pt}
\multirow{2}{*}{\textbf{Method}} & \multicolumn{3}{c|}{\textbf{Lizard}} & \multicolumn{3}{c|}{\textbf{PanNuke}} & \multicolumn{3}{c}{\textbf{Endonuke}}  \\  \cline{2-10}
\multicolumn{1}{c|}{} & Dice & AJI & $F_d$ & Dice & AJI & $F_d$ & Dice & AJI & $F_d$ \\ 
\hline\hline

Baseline &
0.618 & 0.381 & 0.616 & 0.782 & 0.598 &  0.763 & 0.877 & 0.601 & 0.811 \\  
\hline    

Conventional Aug. &
0.675 & 0.423 & 0.646 & 0.816 & 0.641 & 0.791 & 0.891 & 0.624 & 0.823 \\  

\hline
CutOut~\cite{devries2017improved}
& 0.700 & 0.463 & 0.674 & 0.799 & 0.631 & 0.781 & \underline{0.912} & 0.649 & 0.843 \\  
\hline

CutMix~\cite{Yun_2019_ICCV}
& 0.690 & 0.444 & 0.661 & 0.807 & 0.643 & 0.787 & 0.898 & 0.586 & 0.799 \\   
\hline 



SDM~\cite{sdm2022} &
\textbf{0.718} & \textbf{0.487} & \textbf{0.699} & \underline{0.821} & \underline{0.654} & \textbf{0.800} & \textbf{0.914} & \textbf{0.661} &  \underline{0.847} \\  

\hline
Ours &
\underline{0.714} & \underline{0.479} & \underline{0.692} & \textbf{0.822} & \textbf{0.655} & \underline{0.798} & 0.911 & \underline{0.656} & \textbf{0.850} \\ 



\toprule 
\end{tabular}
\label{tab:downstream}
\end{table}

\subsection{Results}

In Table~\ref{tab:label_synthesis}, we demonstrate quantitative results on label synthesis comparing with Yu~\textit{et al.}~\cite{nudiff2023}, Hoogeboom~\textit{et al.}~\cite{hoogeboom2021argmax}, and ablation studies of text conditions. 
First, we use \text{Fr\'{e}chet} Segmentation Distance (FSD)~\cite{FSD} metric. The labels synthesized with Gaussian diffusion model by Yu~\textit{et al.} show a poor FSD score with random-looking noise (see Fig.~\ref{fig:overview}a). 
The categorical diffusion model suggested by Hoogeboom~\textit{et al.} demonstrates a comparable FSD score on the Lizard dataset; however, generated label classes are biased. 
%
%
Similarly, our unconditional model also showed biased generation as depicted in Fig.~\ref{fig:graph}a. 
As we incorporated more structure-related text conditions, our model well utilized the text-image alignment for desired label generation. 
Especially, tissue type text conditioning affects well in PanNuke, the multi-tissue type dataset.

Table~\ref{tab:image_synthesis} shows the image quality evaluation compared to semantic conditional generative methods using \text{Fr\'{e}chet} Inception Distance (FID)~\cite{FID} and Inception Score (IS)~\cite{IS} metrics. 
Our fine-tuned semantic-text conditional LDM achieved the best in FID and second in IS next to SDM. 
However, as shown in Fig.~\ref{fig:overview}
, SDM generated unrealistic color images, which leads to higher IS. 
Moreover, as LDM requires less time for image generation, it can dramatically reduce the time for image generation, up to 33.6$\times$ faster, compared to SDM. 
As a result, the tailored LDM can produce high-quality images that match the semantic and textual conditions, providing a computationally efficient alternative to traditional pixel-based diffusion models for pathology nuclei image synthesis.

We also conducted downstream nuclei segmentation and classification with synthetic data, using patches that were excluded from training the generative model.
We use HoVer-Net~\cite{graham2019hover} as our baseline network for downstream task evaluation. 
We utilize Dice coefficient and Aggregated Jaccard Index (AJI)~\cite{AJI} metrics to quantify segmentation performance, and detection quality $F_d$ for classification. 
In Table~\ref{tab:downstream}, the conventional augmentation improves the baseline performance. 
We generated image-label data from a text prompt that conditions the label generation process and augmented the training set with the same number of patches as SDM's.
For SDM data generation, we used ground truth labels.
%
%
Ours improved baseline with conventional augmentation and comparable score within every metric to SDM. 
Considering the data generation cost of SDM as in Table~\ref{tab:image_synthesis}, our approach is highly promising.

\section{Conclusion}
In this paper, we proposed a novel two-stage multi-class nuclei data augmentation framework, consisting of label synthesis followed by image synthesis.
We showed that the text conditional label synthesis strategy enables manipulation of the synthesized labels.
To the best of our knowledge, this is the first attempt 
to leverage pretrained LDM for nuclei image synthesis aligning with the corresponding label, securing both high-quality sampled images and resource effectiveness.
Lastly, we validate the efficacy of the proposed scheme on downstream tasks, improving the nuclei segmentation and classification performance.

For future work, we plan to explore synthesizing bigger-sized images, such as entire whole slide images, to expand the scalability of generative data augmentation.



\begin{credits}
\subsubsection{\ackname} 
This work was partially supported by the National Research Foundation of Korea (RS-2024-00349697, NRF-2021R1A6A1A13044830), the Institute for Information \& Communications Technology Planning \& Evaluation (IITP-2024-2020-0-01819), the Technology development Program(RS-2024-00437796) funded by the Ministry of SMEs and Startups(MSS, Korea), and 
a Korea University Grant.

\subsubsection{\discintname}
There are no conflicts of interest to declare.

\end{credits}
\bibliographystyle{splncs04}
%
\bibliography{06_ref}

\begin{thebibliography}{10}
\providecommand{\url}[1]{\texttt{#1}}
\providecommand{\urlprefix}{URL }
\providecommand{\doi}[1]{https://doi.org/#1}

\bibitem{alom2022microscopic}
Alom, Z., Asari, V.K., Parwani, A., Taha, T.M.: Microscopic nuclei classification, segmentation, and detection with improved deep convolutional neural networks (dcnn). Diagnostic Pathology  \textbf{17}(1),  1--17 (2022)

\bibitem{FSD}
Bau, D., Zhu, J.Y., Wulff, J., Peebles, W., Strobelt, H., Zhou, B., Torralba, A.: Seeing what a gan cannot generate. In: Proceedings of the IEEE/CVF International Conference on Computer Vision. pp. 4502--4511 (2019)

\bibitem{devries2017improved}
DeVries, T., Taylor, G.W.: Improved regularization of convolutional neural networks with cutout. arXiv preprint arXiv:1708.04552  (2017)

\bibitem{diffusion_beat_gans2021}
Dhariwal, P., Nichol, A.: Diffusion models beat gans on image synthesis. Advances in neural information processing systems  \textbf{34},  8780--8794 (2021)

\bibitem{doan2022gradmix}
Doan, T.N.N., Kim, K., Song, B., Kwak, J.T.: Gradmix for nuclei segmentation and classification in imbalanced pathology image datasets. In: International Conference on Medical Image Computing and Computer-Assisted Intervention. pp. 171--180. Springer (2022)

\bibitem{gamper2020pannuke}
Gamper, J., Koohbanani, N.A., Benes, K., Graham, S., Jahanifar, M., Khurram, S.A., Azam, A., Hewitt, K., Rajpoot, N.: Pannuke dataset extension, insights and baselines. arXiv preprint arXiv:2003.10778  (2020)

\bibitem{gong2021style}
Gong, X., Chen, S., Zhang, B., Doermann, D.: Style consistent image generation for nuclei instance segmentation. In: Proceedings of the IEEE/CVF winter conference on applications of computer vision. pp. 3994--4003 (2021)

\bibitem{graham2021lizard}
Graham, S., Jahanifar, M., Azam, A., Nimir, M., Tsang, Y.W., Dodd, K., Hero, E., Sahota, H., Tank, A., Benes, K., et~al.: Lizard: a large-scale dataset for colonic nuclear instance segmentation and classification. In: Proceedings of the IEEE/CVF International Conference on Computer Vision. pp. 684--693 (2021)

\bibitem{graham2019hover}
Graham, S., Vu, Q.D., Raza, S.E.A., Azam, A., Tsang, Y.W., Kwak, J.T., Rajpoot, N.: Hover-net: Simultaneous segmentation and classification of nuclei in multi-tissue histology images. Medical image analysis  \textbf{58},  101563 (2019)

\bibitem{FID}
Heusel, M., Ramsauer, H., Unterthiner, T., Nessler, B., Hochreiter, S.: Gans trained by a two time-scale update rule converge to a local nash equilibrium. Advances in neural information processing systems  \textbf{30} (2017)

\bibitem{ddpm2020}
Ho, J., Jain, A., Abbeel, P.: Denoising diffusion probabilistic models. Advances in neural information processing systems  \textbf{33},  6840--6851 (2020)

\bibitem{hoogeboom2021argmax}
Hoogeboom, E., Nielsen, D., Jaini, P., Forr{\'e}, P., Welling, M.: Argmax flows and multinomial diffusion: Learning categorical distributions. Advances in Neural Information Processing Systems  \textbf{34},  12454--12465 (2021)

\bibitem{ilyas2022tsfd}
Ilyas, T., Mannan, Z.I., Khan, A., Azam, S., Kim, H., De~Boer, F.: Tsfd-net: Tissue specific feature distillation network for nuclei segmentation and classification. Neural Networks  \textbf{151},  1--15 (2022)

\bibitem{AJI}
Kumar, N., Verma, R., Sharma, S., Bhargava, S., Vahadane, A., Sethi, A.: A dataset and a technique for generalized nuclear segmentation for computational pathology. IEEE transactions on medical imaging  \textbf{36}(7),  1550--1560 (2017)

\bibitem{naumov2022endonuke}
Naumov, A., Ushakov, E., Ivanov, A., Midiber, K., Khovanskaya, T., Konyukova, A., Vishnyakova, P., Nora, S., Mikhaleva, L., Fatkhudinov, T., et~al.: Endonuke: Nuclei detection dataset for estrogen and progesterone stained ihc endometrium scans. Data  \textbf{7}(6), ~75 (2022)

\bibitem{nichol2021improved}
Nichol, A.Q., Dhariwal, P.: Improved denoising diffusion probabilistic models. In: International Conference on Machine Learning. pp. 8162--8171. PMLR (2021)

\bibitem{diffmix2023}
Oh, H.J., Jeong, W.K.: Diffmix: Diffusion model-based data synthesis for nuclei segmentation and classification in imbalanced pathology image datasets. In: Medical Image Computing and Computer Assisted Intervention -- MICCAI 2023. pp. 337--345. Springer (2023)

\bibitem{park_gcdp2023}
Park, M., Yun, J., Choi, S., Choo, J.: Learning to generate semantic layouts for higher text-image correspondence in text-to-image synthesis. In: Proceedings of the IEEE/CVF International Conference on Computer Vision. pp. 7591--7600 (2023)

\bibitem{spade2019}
Park, T., Liu, M.Y., Wang, T.C., Zhu, J.Y.: Semantic image synthesis with spatially-adaptive normalization. In: Proceedings of the IEEE/CVF conference on computer vision and pattern recognition. pp. 2337--2346 (2019)

\bibitem{rombach2022_stablediffusion}
Rombach, R., Blattmann, A., Lorenz, D., Esser, P., Ommer, B.: High-resolution image synthesis with latent diffusion models. In: Proceedings of the IEEE/CVF conference on computer vision and pattern recognition. pp. 10684--10695 (2022)

\bibitem{saharia2205photorealistic}
Saharia, C., Chan, W., Saxena, S., Li, L., Whang, J., Denton, E.L., Ghasemipour, K., Gontijo~Lopes, R., Karagol~Ayan, B., Salimans, T., et~al.: Photorealistic text-to-image diffusion models with deep language understanding. Advances in neural information processing systems  \textbf{35},  36479--36494 (2022)

\bibitem{IS}
Salimans, T., Goodfellow, I., Zaremba, W., Cheung, V., Radford, A., Chen, X.: Improved techniques for training gans. Advances in neural information processing systems  \textbf{29} (2016)

\bibitem{nasdm2023}
Shrivastava, A., Fletcher, P.T.: Nasdm: Nuclei-aware semantic histopathology image generation using diffusion models. In: Medical Image Computing and Computer Assisted Intervention -- MICCAI 2023. pp. 786--796. Springer (2023)

\bibitem{song2020denoising_ddim}
Song, J., Meng, C., Ermon, S.: Denoising diffusion implicit models. arXiv preprint arXiv:2010.02502  (2020)

\bibitem{wang2023sian}
Wang, H., Xian, M., Vakanski, A., Shareef, B.: Sian: style-guided instance-adaptive normalization for multi-organ histopathology image synthesis. In: 2023 IEEE 20th International Symposium on Biomedical Imaging (ISBI). pp.~1--5. IEEE (2023)

\bibitem{pix2pixHD2018}
Wang, T.C., Liu, M.Y., Zhu, J.Y., Tao, A., Kautz, J., Catanzaro, B.: High-resolution image synthesis and semantic manipulation with conditional gans. In: Proceedings of the IEEE conference on computer vision and pattern recognition. pp. 8798--8807 (2018)

\bibitem{sdm2022}
Wang, W., Bao, J., Zhou, W., Chen, D., Chen, D., Yuan, L., Li, H.: Semantic image synthesis via diffusion models. arXiv preprint arXiv:2207.00050  (2022)

\bibitem{yang2006nuclei}
Yang, X., Li, H., Zhou, X.: Nuclei segmentation using marker-controlled watershed, tracking using mean-shift, and kalman filter in time-lapse microscopy. IEEE Transactions on Circuits and Systems I: Regular Papers  \textbf{53}(11),  2405--2414 (2006)

\bibitem{pathldm2024}
Yellapragada, S., Graikos, A., Prasanna, P., Kurc, T., Saltz, J., Samaras, D.: Pathldm: Text conditioned latent diffusion model for histopathology. In: Proceedings of the IEEE/CVF Winter Conference on Applications of Computer Vision. pp. 5182--5191 (2024)

\bibitem{nudiff2023}
Yu, X., Li, G., Lou, W., Liu, S., Wan, X., Chen, Y., Li, H.: Diffusion-based data augmentation for nuclei image segmentation. In: International Conference on Medical Image Computing and Computer-Assisted Intervention. pp. 592--602. Springer (2023)

\bibitem{Yun_2019_ICCV}
Yun, S., Han, D., Oh, S.J., Chun, S., Choe, J., Yoo, Y.: Cutmix: Regularization strategy to train strong classifiers with localizable features. In: Proceedings of the IEEE/CVF International Conference on Computer Vision (ICCV) (October 2019)

\bibitem{zhang2023_controlnet}
Zhang, L., Rao, A., Agrawala, M.: Adding conditional control to text-to-image diffusion models. In: Proceedings of the IEEE/CVF International Conference on Computer Vision. pp. 3836--3847 (2023)

\end{thebibliography}




\end{document}